\documentclass[letterpaper, 10 pt, conference]{ieeeconf}  

\IEEEoverridecommandlockouts                              

\usepackage{graphics} 
\usepackage{epsfig} 
\usepackage{times} 
\usepackage{amsmath} 
\usepackage{amssymb}  

\usepackage{graphicx}
\usepackage{tabularx}
\usepackage{caption} 
\usepackage{wrapfig}
\usepackage{romannum}
\usepackage{url}
\usepackage{bm}
\usepackage[table,xcdraw]{xcolor}
\usepackage{tabularray}
\usepackage[colorlinks=true]{hyperref}
\usepackage{cite}

\usepackage{multirow}
\usepackage{booktabs}
\usepackage{pifont}
\usepackage{arydshln}

\usepackage{bbm} 
\usepackage[table]{xcolor} 
\usepackage{multirow} 
\newcommand{\bth}{\boldsymbol{\theta}}
\newcommand{\bp}{\mathbf{p}}
\newcommand{\bpi}{\mathbf{p}_i}
\newcommand{\byi}{\mathbf{y}_i}

\definecolor{puddle}{rgb}{0.0, 0.0, 1.0}
\definecolor{object}{rgb}{0.8, 0.6, 1.0}
\definecolor{paved}{rgb}{1.0, 1.0, 0.0}
\definecolor{unpaved}{rgb}{1.0, 0.6, 0.8}
\definecolor{dirt}{rgb}{0.6, 0.298039, 0.0}
\definecolor{grass}{rgb}{0.435294, 1.0, 0.290196}
\definecolor{vegetation}{rgb}{0.0, 0.4, 0.0}

\newcommand\semcolor[1][black]{\fcolorbox{black}{#1}{\rule{0mm}{1mm}\rule{1mm}{0mm}}}

\newcommand\semcolorbf[2]{\textbf{\textcolor{#1}{#2}}}

\newcommand{\ourdataName}{OffRoad}

\newcommand{\ourdata}{\ourdataName}
\newcommand{\ourdatabf}{\textbf{\ourdataName}}

\newcommand{\ourdataP}{\ourdataName-P}
\newcommand{\ourdataPbf}{\textbf{\ourdataName-P}}

\newcommand{\calL}{\mathcal{L}}

\newcommand{\Dkl}{D_{\text{KL}}}
\newcommand{\dataset}{\mathcal{D} = \{x_i, \byi\}^N_{i=1}}
\newcommand{\datasetSem}{\mathcal{D}_{\text{sem}} = \{x_i, \byi\}^N_{i=1}}
\newcommand{\datasetp}{\mathcal{D'} = \{x_i, \bp_i, u_i\}^N_{i=1}}
\newcommand{\Uthres}{\mathrm{U}_{\text{thr}}}
\newcommand{\softmax}{\text{SoftMax}(\cdot)}

\newcommand{\rellis}{RELLIS-3D }

\newcommand{\dol}{\frac{d}{l}}
\newcommand{\tpk}{\text{TP}_c}
\newcommand{\fpk}{\text{FP}_c}
\newcommand{\fnk}{\text{FN}_c}

\newcommand{\Tref}[1]{Table~\ref{#1}}
\newcommand{\Fref}[1]{Fig.~\ref{#1}}

\newcommand{\Sref}[1]{Section~\ref{#1}}

\DeclareCaptionFont{mysize}{\fontsize{8}{9.6}\selectfont}
\title{\LARGE \bf
Evidential Semantic Mapping in Off-road Environments with Uncertainty-aware Bayesian Kernel Inference
}

\author{Junyoung Kim$^{*}$, Junwon Seo$^{*}$, Jihong Min%
\thanks{This work was supported by the Agency for Defense Development Grant funded by the Korean Government in 2024.}
\thanks{Junyoung Kim, Junwon Seo, and Jihong Min are with the Agency for Defense Development, Republic of Korea {\tt\footnotesize \{junyoung.kimv, junwon.vision, happymin77\}@gmail.com}}%
\thanks{$^{*}$These authors contributed equally to this work.}
\thanks{Our project website can be found at \href{https://bit.ly/EvSemMap}{\tt\footnotesize https://bit.ly/EvSemMap}}%
}

\begin{document}

\maketitle

\begin{abstract}
Robotic mapping with Bayesian Kernel Inference~(BKI) has shown promise in creating semantic maps by effectively leveraging local spatial information. However, existing semantic mapping methods face challenges in constructing reliable maps in unstructured outdoor scenarios due to unreliable semantic predictions. To address this issue, we propose an evidential semantic mapping, which can enhance reliability in perceptually challenging off-road environments. We integrate Evidential Deep Learning into the semantic segmentation network to obtain the uncertainty estimate of semantic prediction. Subsequently, this semantic uncertainty is incorporated into an uncertainty-aware BKI, tailored to prioritize more confident semantic predictions when accumulating semantic information. By adaptively handling semantic uncertainties, the proposed framework constructs robust representations of the surroundings even in previously unseen environments. Comprehensive experiments across various off-road datasets demonstrate that our framework enhances accuracy and robustness, consistently outperforming existing methods in scenes with high perceptual uncertainties.
\end{abstract}
\section{INTRODUCTION}
\newcommand{\citeOccupancyMap}{\cite{131_OccupancyMap_elfes1989using, 6_OctoMap_hornung2013octomap}}
\newcommand{\citeSparsity}{\cite{50_GPOctoMap_wang2016fast}}
\newcommand{\citeSemanticMap}{\cite{94_kim20133d, 103_valentin2013mesh, 95_SemanticOctree_sengupta2015semantic, 97_yang2017semantic, 98_DA-RNN_xiang2017rnn, 96_SemanticFusion_mccormac2017semanticfusion, 109_paz2020probabilistic, 23_SSMI_asgharivaskasi2023semantic, 26_morilla2023robust, 106_FusionOverconfidence_marques2023overconfidence}}
\newcommand{\citeSemanticMapCompact}{\cite{94_kim20133d, 103_valentin2013mesh, 95_SemanticOctree_sengupta2015semantic, 109_paz2020probabilistic, 23_SSMI_asgharivaskasi2023semantic, 26_morilla2023robust, 106_FusionOverconfidence_marques2023overconfidence}}
\newcommand{\citeVoxelSemanticMap}{\cite{94_kim20133d, 95_SemanticOctree_sengupta2015semantic, 97_yang2017semantic, 98_DA-RNN_xiang2017rnn, 23_SSMI_asgharivaskasi2023semantic, 26_morilla2023robust, 106_FusionOverconfidence_marques2023overconfidence}}
\newcommand{\citeContinuousMap}{\cite{50_GPOctoMap_wang2016fast, 49_BGKOctoMap_doherty2017bayesian}}
\newcommand{\citeCRFSemanticMap}{\cite{92_sengupta2013urban, 103_valentin2013mesh, 94_kim20133d, 101_kundu2014joint, 95_SemanticOctree_sengupta2015semantic, 104_vineet2015incremental}}
\newcommand{\citeDNNbasedSemanticMap}{\cite{96_SemanticFusion_mccormac2017semanticfusion, 97_yang2017semantic, 98_DA-RNN_xiang2017rnn, 99_RecurrentOctoMap_sun2018recurrent, 105_maturana2018real}}

Robotic mapping, which aims to construct an explicit environmental model using sensor measurements, is an essential component in various robotic systems. Conventional approaches, such as occupancy grid mapping~\citeOccupancyMap, construct 3D voxel maps by estimating the occupancy state of each voxel using sensor measurements intersecting with the voxel. However, due to their independent voxel assumptions, these approaches often produce discontinuous and sparse maps, especially when dealing with sparse sensor data like LiDAR~\citeSparsity. Consequently, recent works have focused on enhancing the geometric completeness of occupancy estimation by leveraging local spatial information~\citeContinuousMap. Specifically, the adoption of Bayesian Kernel Inference~(BKI)~\cite{52_BKI_vega2014nonparametric} has enabled the efficient utilization of neighboring measurements, resulting in more reliable continuous maps~\cite{49_BGKOctoMap_doherty2017bayesian}.

Semantic mapping has significantly improved the ability of autonomous systems to comprehend their surroundings across various robotic applications~\citeSemanticMapCompact. By estimating voxel-wise semantic states, it provides a richer understanding of the environments. Building upon the advancements in continuous mapping, S-BKI~\cite{51_S-BKI_gan2020bayesian} extended BKI into $3$D continuous semantic mapping. Notably, recent semantic mapping approaches depend on the inference of Deep Neural Networks~(DNNs)~\citeDNNbasedSemanticMap. These networks derive semantic predictions from sensor measurements, which are then integrated into the mapping process. As a result, the performance of semantic mapping is highly dependent on the accuracy of semantic predictions generated by DNNs.

\begin{figure}[t]
\centering
\includegraphics[width=1.0\linewidth]{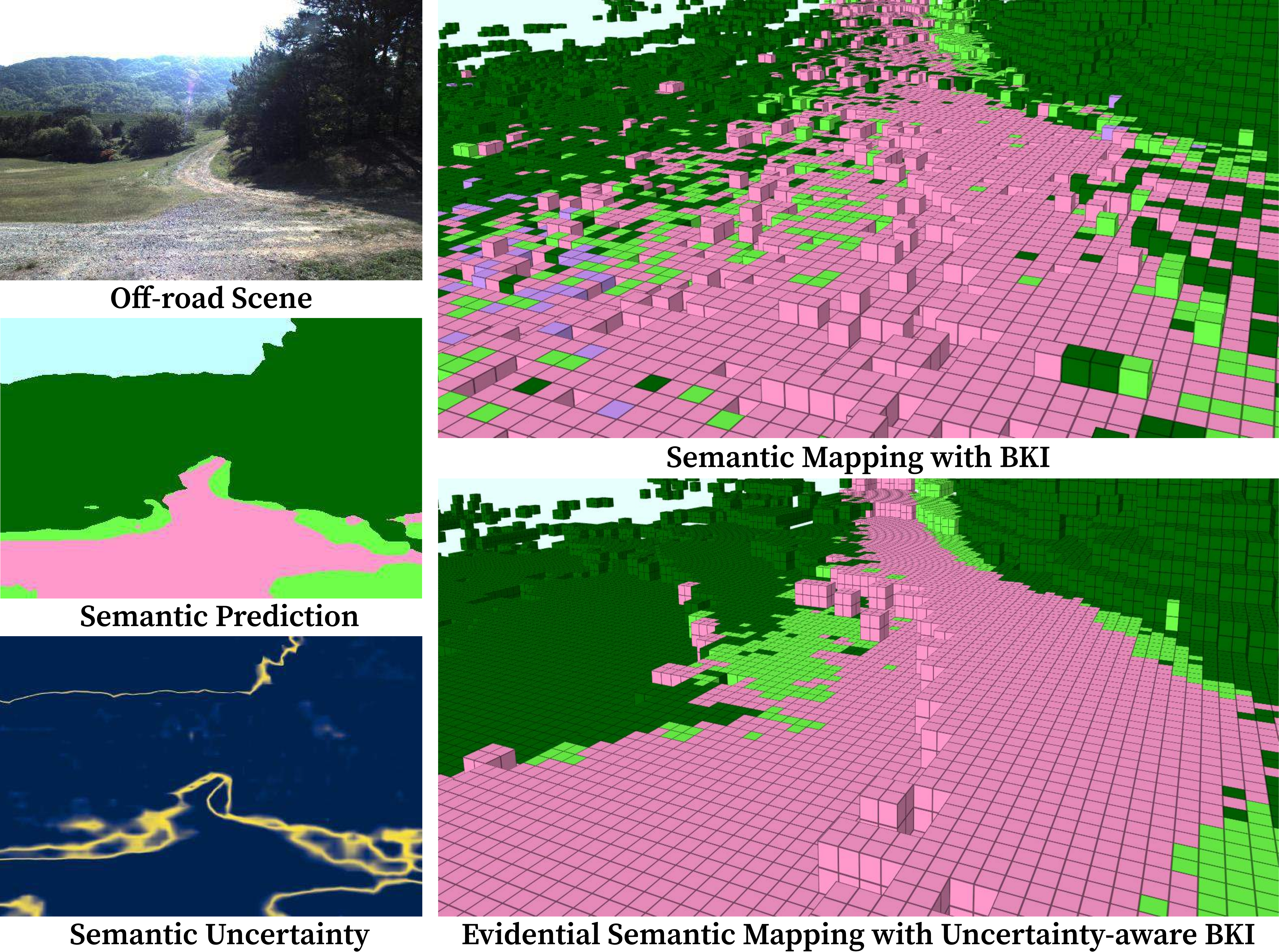}
\caption{In perceptually challenging unstructured outdoor environments, considering semantic uncertainty from the segmentation network is beneficial to produce more accurate semantic maps by prioritizing confident predictions. By incorporating the semantic uncertainty of the network, our uncertainty-aware semantic BKI mapping framework can produce reliable and accurate 3D semantic maps in off-road scenes.}
\label{fig:concept}
\vspace{-0.2in}
\end{figure}

However, the predictions from DNNs often suffer from unreliability~\cite{107_guo2017calibration, 83_ModernReliability_de2023reliability}, compromising the mapping performance, especially in off-road settings. The diverse appearances and unstructured nature of off-road environments, combined with illumination changes that increase intra-class variation, lead to unreliable operation of DNNs~\cite{86_OFFROAD_jin2021memory, seo2023learning}. Moreover, the common approach of compressing probabilistic predictions from DNNs into one-hot vectors~\cite{51_S-BKI_gan2020bayesian, 88_SEE-CSOM_deng2023see, 47_ConvBKI2_wilson2023convbki} results in the loss of valuable information, further exacerbating mapping performance. Consequently, semantic mapping in perceptually challenging off-road environments frequently leads to unreliable outputs. To address this challenge, it is crucial to consider the uncertainty of semantic predictions from DNNs to construct semantic maps robustly, underscoring the necessity for advanced, uncertainty-aware mapping techniques, as illustrated in \Fref{fig:concept}.

In this work, we propose an uncertainty-aware semantic BKI mapping framework for robust deployments in off-road environments with high perceptual uncertainties. To achieve this goal, we incorporate the uncertainty of semantic predictions into the mapping process, thereby improving the reliability and accuracy of the map. To estimate the semantic uncertainty, we adopt Evidential Deep Learning~\cite{10_EDL_sensoy2018evidential}, known for its reliable uncertainty estimates. Then, probabilistic semantic predictions are seamlessly integrated into our mapping process, enabling precise representations of highly uncertain environments. Furthermore, the mapping process is adapted to incorporate semantic uncertainty by not only filtering out uncertain predictions but also reinforcing the effect of confident predictions. Through validation with off-road datasets, including \rellis and our own off-road datasets, we verify that our method enhances the accuracy and reliability of semantic mapping compared to existing methods.

\section{RELATED WORK}
\subsection{3D Semantic Mapping}
Among various 3D geometric representations such as Gaussian~\cite{4_GMMap_li2024gmmap}, surfel~\cite{96_SemanticFusion_mccormac2017semanticfusion}, and mesh~\cite{103_valentin2013mesh}, voxel-based 3D model is commonly used in semantic mapping~\citeVoxelSemanticMap. Since traditional voxel-based mapping techniques assume the statistical independence of each voxel, each sensor measurement is used to update only a restricted subset of voxels. As a result, sparse LiDAR measurements lead to the construction of discontinuous maps~\cite{50_GPOctoMap_wang2016fast}. Therefore, methods have been proposed to relax the voxel independence assumption by incorporating local spatial information, demonstrating improved performance~\citeContinuousMap. 

GPOctoMap~\cite{50_GPOctoMap_wang2016fast} introduces Gaussian Process~(GP) to account for local spatial correlations, relaxing the voxel independence assumption. However, the computational cost $\mathcal{O}(n^3)$ with $n$ being the number of sensor measurements makes it prohibitive for real-time applications. To address this gap, Bayesian Kernel Inference~(BKI)~\cite{52_BKI_vega2014nonparametric} has been introduced as an efficient approximation to GP~\cite{49_BGKOctoMap_doherty2017bayesian}. This significantly reduces the computational complexity to $\mathcal{O}(\log n)$, facilitating real-time updates and efficient integration of local spatial information into the mapping process. This efficient BKI was then adopted in S-BKI~\cite{51_S-BKI_gan2020bayesian} for semantic mapping, extending the capability to include semantic information within the geometric model.

While traditional semantic mapping systems have utilized Conditional Random Field~(CRF)~\citeCRFSemanticMap, subsequent improvements have increasingly favored the usage of Deep Neural Networks~(DNNs) to incorporate semantic information into maps~\citeDNNbasedSemanticMap. The semantic segmentation results obtained from DNNs at each time step are commonly utilized for the incremental update of the semantic map. After S-BKI~\cite{51_S-BKI_gan2020bayesian} introduces the BKI-based continuous mapping into the semantic mapping field, various approaches are proposed to improve the accuracy of continuous semantic mapping. ConvBKI~\cite{47_ConvBKI2_wilson2023convbki} proposes a trainable version of S-BKI using depth-wise convolution, encouraging the kernel to learn class-specific geometry. Meanwhile, SEE-CSOM~\cite{88_SEE-CSOM_deng2023see} is proposed to address the challenges of overinflation and inefficiency by considering the label inconsistency of the voxels. However, the uncertainty of semantic predictions is yet to be introduced in continuous semantic mapping.

\subsection{Uncertainty Estimation of Deep Neural Networks}
\newcommand{\citeEDLVariousField}{\cite{70_USNet_chang2022fast, 72_TMCJournal_han2022trusted, 13_EvPSNet_sirohi2023uncertainty, 120_EVORA_cai2023evora}}

Uncertainty estimation~\cite{107_guo2017calibration, 83_ModernReliability_de2023reliability}, considered an important aspect of deploying DNNs in safety-critical applications, is a process of estimating the uncertainty of the predictions from DNNs. The most widely used methods are sampling-based approaches~\cite{118_BNNs_jospin2022hands}, requiring multiple inferences for each input. For example, Monte Carlo~(MC) dropout~\cite{117_MCDropout_gal2016dropout} performs inference multiple times with a stochastic dropout layer, and Deep Ensemble~\cite{110_DeepEnsemble_lakshminarayanan2017simple} uses an ensemble of networks to obtain multiple predictions to estimate uncertainty. Therefore, they are unsuitable for resource-constrained real-time applications~\cite{kim2023bridging}. Thus, methods that enable uncertainty estimation with a single forward pass, such as Evidential Deep Learning~(EDL)~\cite{10_EDL_sensoy2018evidential}, have been proposed.

EDL framework estimates uncertainty by modeling a second-order Dirichlet distribution based on the Dempster-Shafer Theory of Evidence~\cite{129_DST_yager2008classic}. The network is trained to compute parameters for the Dirichlet distribution from which the uncertainty of prediction is calculated. Since it can be easily applied to existing DNNs, it has been widely adopted in various fields for uncertainty estimation~\citeEDLVariousField. 

\section{METHODS}
\begin{figure*}[t]
\begin{center}
\includegraphics[width=1.0\linewidth]{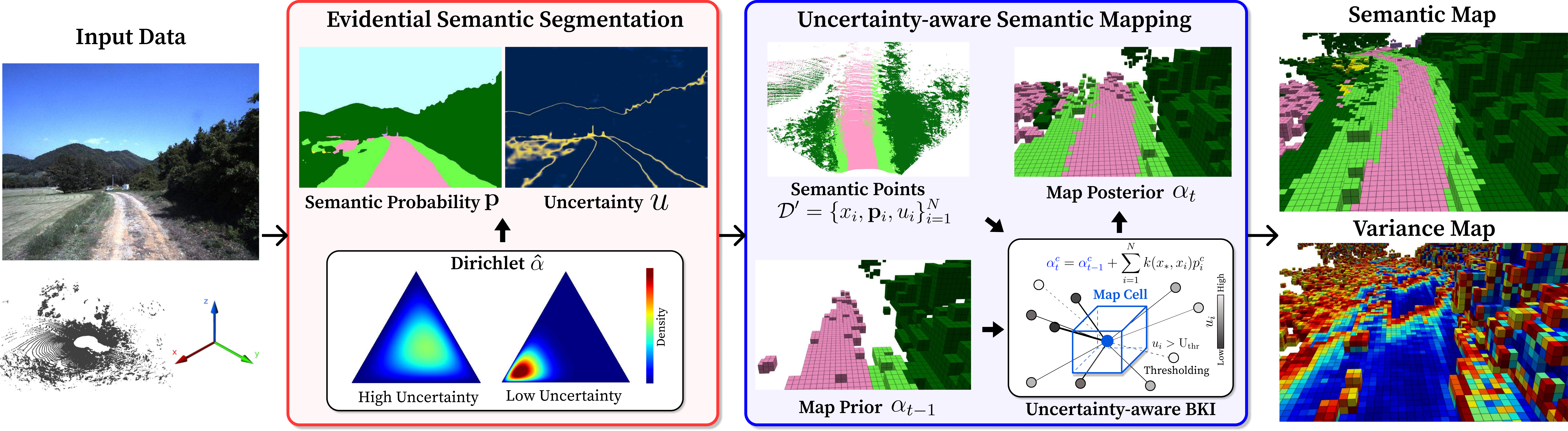}
\end{center}
\caption{Overview pipeline of our uncertainty-aware semantic BKI framework. With an evidential segmentation network trained by EDL, input data is processed to derive continuous semantic probability and uncertainty. These 3D semantic points are then integrated into the semantic map through Bayesian updates using the uncertainty-aware BKI, resulting in a dependable semantic map and variance map in uncertain off-road environments.}
\label{fig:framework}
\vspace{-0.2in}
\end{figure*}

In this section, we present our uncertainty-aware semantic mapping framework. First, we provide an overview of the semantic BKI mapping process. Then, we incorporate the EDL framework into DNNs for semantic segmentation to estimate semantic uncertainty. Lastly, we introduce our uncertainty-aware BKI framework, which enhances the performance of semantic mapping by leveraging the predictive uncertainty of semantic predictions. Our entire framework is illustrated in \Fref{fig:framework}.

\subsection{Semantic Bayesian Kernel Inference}
In semantic BKI frameworks, the mapping objective is to estimate the probability distribution $\bth_* = [\theta_*^1, ..., \theta_*^C]$ across $C$ semantic classes for arbitrary query point $x_*$, where $\sum^C_{c=1} \theta_*^c = 1$ and $\theta_*^c \in [0,1]$. The $\bth_*$ is then used to predict semantic label $\mathbf{y}_*$ based on the Categorical distribution $p(\mathbf{y}_*|\bth_*)$. Thus, the semantic points $\dataset$ is utilized to estimate the posterior distribution $p(\bth_* | x_*, \mathcal{D})$, where $x_i$ is the coordinates of the $i$-th observation and $\byi$ represents the corresponding semantic label. In practice, the posterior distribution is estimated only for the center of voxels during the mapping for efficiency, though queries on arbitrary locations can still be processed.

Initially, Bayes' rule is applied to derive the posterior in closed form equation:
\begin{equation}
\begin{aligned}\label{BKI_1_Bayes}
    p(\bth_*|x_*, \mathcal D)  
        &\propto p(\mathcal{D} | \bth_*, x_*) p(\bth_* | x_*)\\
        &\propto 
            \Big[ \prod^N_{i=1} 
                \underbrace {\strut p(\byi | x_i, \bth_*, x_*) }_{\text{extended likelihood}} 
            \Big]
            \underbrace{\strut p(\bth_* | x_*)}_{\text{prior}} .
\end{aligned}
\end{equation}
Then, BKI \cite{52_BKI_vega2014nonparametric} is applied as it provides the relationship between the extended likelihood and the likelihood. In \cite{52_BKI_vega2014nonparametric}, the maximum entropy distribution $g$ satisfying $\Dkl (g || f) \le \rho(x_*, x)$ is proven to be $g(y) \propto f(y) ^ {k(x_*,x)}$ with a kernel function $k : \mathcal X \times \mathcal X \rightarrow [0, 1]$, where $\Dkl(\cdot || \cdot)$ is the Kullback-Leibler divergence and $\rho(\cdot, \cdot)$ is a smoothness bound. Substituting $g$ for the extended likelihood and $f$ for the likelihood, we can obtain the following proportional relationship:
\begin{align}\label{BKI_2_VegaBrown}
    p(\byi | x_i, \bth_*, x_*) \propto p(\byi | \bth_*)^{k(x_*, x_i)} .
\end{align}
By combining \eqref{BKI_1_Bayes} and \eqref{BKI_2_VegaBrown}, we have:
\begin{align}\label{BKI_3_final}
    p(\bth_*|x_*, \mathcal D)  \propto \Big[ \prod^N_{i=1} 
                p(\byi | \bth_*)^{k(x_*, x_i)}
            \Big] p(\bth_* | x_*) .
\end{align}

As $\byi$ represents the one-hot encoded semantic label and the Categorical likelihood $p(y_i|\bth_*)$ is used in the frameworks, a Dirichlet prior is adopted to enable incremental Bayesian inference since it is the conjugate prior distribution. An initial Dirichlet prior $Dir(C, \boldsymbol{\alpha}_0)$ over $\bth_*$ is parameterized by hyperparameters $\boldsymbol{\alpha}_0 = [\alpha_0^1, ..., \alpha_0^C]$. Applying the definition of the prior and the likelihood, \eqref{BKI_3_final} becomes:
\begin{equation}
\begin{aligned}
    p(\theta_*|x_*, \mathcal D) &\propto \Big[ \prod^N_{i=1} \bigl(
        \prod^C_{c=1} (\theta_*^c)^{y_i^c} 
    \bigr)^{k(x_*, x_i)} \Big] \prod^C_{c=1} (\theta_*^c)^{\alpha_0^c - 1} \\
    &\propto \prod^C_{c=1} (\theta_*^c)^{\alpha_0^c + \sum^N_{i=1} k(x_*, x_i)y_i^c - 1} .
\end{aligned}
\end{equation}

Consequently, a Dirichlet posterior $Dir(C, \boldsymbol{\alpha}_{t})$ at time $t$ can be calculated based on the posterior at time $t-1$. This update process for the map cell can be recursively executed as:
\begin{equation}
\begin{aligned}\label{BKI_final_update}
    \alpha^c_{t} = {\alpha_{t-1}^c + \sum^N_{i=1} k(x_*, x_i)y^c_i} ,
\end{aligned}
\end{equation}
and the expectation and variance of the posterior are calculated as:
\begin{equation}
\begin{aligned}\label{BKI_final_variance}
    S_{t} = \sum^C_{c=1} \alpha_{t}^c, \ \
    \mathbb{E}[\alpha_{t}^c] = \frac{\alpha_{t}^c}{S_{t}}, \ \ 
    \mathrm{Var}[\alpha_{t}^c] = \frac{\alpha_{t}^c (S_{t} - \alpha_{t}^c)}{S_{t}^2(S_{t} + 1)} .
\end{aligned}
\end{equation}
It is worth noting that the variance of the predicted label $\mathrm{Var}[\alpha_t^\psi]$ is utilized as an uncertainty measure of the map cell in \cite{51_S-BKI_gan2020bayesian, 47_ConvBKI2_wilson2023convbki}, where $\psi = \mathrm{argmax}_c \mathbb{E}[\alpha_t^c]$.

For the kernel function $k(x_*, x_i)$ in \eqref{BKI_final_update}, it is desired to prioritize information from nearby points and ignore information from the points that are too distant. Among the functions with these characteristics, the sparse kernel \cite{66_SparseKernel_melkumyan2009sparse} is commonly used to ensure smoothness and reduce computational complexity \cite{49_BGKOctoMap_doherty2017bayesian, 51_S-BKI_gan2020bayesian, 88_SEE-CSOM_deng2023see}:
\begin{equation}\label{BKI_5_reparam}
\begin{aligned}
    k(x_*, x_i) &= k'(d, l, \sigma_0) \\
                &=\underset{d < l}{\mathbbm{1}} \: \sigma_0 \Big[ \frac{2 + \cos (2\pi \dol)}{3} (1 - \dol) + \frac{1}{2\pi} \sin (2\pi \dol) \Big] ,
\end{aligned}
\end{equation}
where $\mathbbm{1}$ is the indicator function, $d = ||x_* - x_i||$ is the distance between $x_*$ and $x_i$, $l > 0$ is the length-scale hyperparameter, and $\sigma_0$ is a kernel scaling hyperparameter. As the value of the sparse kernel decreases smoothly as the distance increases up to $l$, the surrounding size of space considered by the BKI framework can be scaled by adjusting~$l$.

\subsection{Evidential Semantic Segmentation}\label{METHOD_EDL}
To utilize the semantic uncertainty of predictions in the mapping process, we apply the EDL framework~\cite{10_EDL_sensoy2018evidential} into existing segmentation networks to estimate its uncertainty reliably. The output of the networks for $i$-th input is regarded as an evidence vector $\mathbf e_i = [e_i^1, ..., e_i^C]$ by replacing the last $\softmax$ of the network into $\exp(\cdot)$~\cite{74_RED_pandey2023learn}. 

From the evidence vector $\mathbf e_i$, the uncertainty $u_i$ is estimated as $u_i= C / \hat{S}_i$, where $\hat{S}_i = \sum^C_{c=1}(e_i^c+1)$, enabling deterministic uncertainty estimation with a single-pass operation. Then, the semantic probability distribution $\bpi = [p_i^1, ..., p_i^C]$ can be modeled as $p_i^c = \alpha_i^c / \hat{S}_i$ by the Dirichlet distribution with concentration parameters of EDL $\boldsymbol{\hat{\alpha}}_i = [\hat{\alpha}_i^1, ..., \hat{\alpha}_i^C]$, where $\hat{\alpha}_i^c = e_i^c + 1$. 

We adopt the loss from~\cite{10_EDL_sensoy2018evidential, 13_EvPSNet_sirohi2023uncertainty} to train the evidential segmentation network with EDL. The following evidential loss encourages collecting semantic evidence of its corresponding one-hot label $\byi$ from the training dataset $\datasetSem$ for each pixel $x_i$:
\begin{equation}\label{EDL_evidential_loss}
\begin{aligned}
    \calL_{\text{EDL}} (x_i, y_i) = \sum^C_{c=1} y_i^c (\log(\hat{S}_i) - \log(\hat{\alpha}_i^c)).
\end{aligned}
\end{equation}
While the EDL loss in \eqref{EDL_evidential_loss} can theoretically encourage minimizing $\hat{S}_i$ and maximizing the concentration parameter of ground-truth class $\hat{\alpha}_i^c$, an additional regularization term is commonly utilized to penalize incorrect evidence on non-ground-truth class. The most widely used KL regularization term~\cite{10_EDL_sensoy2018evidential} is adopted as follows:
\begin{align}
    \calL_{\text{reg}} (x_i) =  \Dkl\bigl[\text{Dir}\left(\bpi | \boldsymbol{\tilde{\alpha}}_i \right)\ ||\ \text{Dir}\left(\bpi | \langle1, ..., 1 \rangle \right) \bigr] ,
\end{align}
where $\boldsymbol{\tilde{\alpha}}_i = [\tilde{\alpha}_i^1, ..., \tilde{\alpha}_i^C]$ is the Dirichlet parameters after the removal of correct evidence from $\boldsymbol{\hat{\alpha}_i}$ to penalize the misleading evidence only:
\begin{equation}
\begin{aligned}
    \tilde{\alpha}_i^c = 
        &\begin{cases}
            1 &\text{if } y_i^c = 1 \\
            \hat{\alpha}_i^c &\text{if } y_i^c = 0 \\
        \end{cases} .
\end{aligned}
\end{equation}
The overall loss for our EDL segmentation network is
\begin{align}\label{EDL_OVERALL_LOSS}
    \calL = \calL_{\text{EDL}} +  \lambda_{\text{kl}} \calL_{\text{reg}} , 
\end{align}
where $\lambda_{\text{kl}}$ is the hyperparameter.

Once the evidential model is trained to minimize the loss in \eqref{EDL_OVERALL_LOSS}, the probability $\bpi$ and its uncertainty $u_i$ is reliably estimated for each input $x_i$. By incorporating these estimates into the mapping process, we can improve the performance of semantic mapping even in perceptually challenging environments such as off-road settings.

\subsection{Uncertainty-aware Mapping Extension}\label{METHOD_UAM}
Since the knowledge from the evidential semantic segmentation networks cannot be fully utilized through a squeezed one-hot prediction $\byi$, we generalize the semantic points $\dataset$ to $\datasetp$, where $\bpi$ and $u_i$ are provided from the evidential semantic segmentation network. Accordingly, the likelihood $p(\bpi | \bth_*)$ is also generalized as the Continuous Categorical \cite{89_gordon2020continuous} by
\begin{align}
    p(\bp_i | \bth_*) \propto \prod^C_{c=1} (\theta_*^c)^{p_i^c} .
\end{align}
As the Dirichlet distribution is also the conjugate prior of the Continuous Categorical, we can similarly derive the posterior distribution from \eqref{BKI_3_final}:
\begin{equation}
\begin{aligned}
    p(\bth_*|x_*, \mathcal D') &\propto \Big[ \prod^N_{i=1} \bigl(
        \prod^C_{c=1} (\theta_*^c)^{p_i^c}
    \bigr)^{k(x_*, x_i)} \Big] \prod^C_{c=1} (\theta_*^c)^{\alpha_0^c - 1} \\
    &\propto \prod^C_{c=1} (\theta_*^c)^{\alpha_0^c + \sum^N_{i=1} k(x_*, x_i)p_i^c - 1} .
\end{aligned}
\end{equation}
Hence, the formula for the recursive update of the Dirichlet posterior parameters in \eqref{BKI_final_update} becomes
\begin{equation}
\begin{aligned}\label{uBKI_final_update}
    \alpha^c_t = {\alpha_{t-1}^c + \sum^N_{i=1} k(x_*, x_i)p^c_i}.
\end{aligned}
\end{equation}
Building upon the extension with the Continuous Categorical, the uncertain off-road environments can be accurately represented through semantic points compared to utilizing one-hot squeezed segmentation results, thereby enhancing semantic mapping performances.

Furthermore, we devise an uncertainty-aware adaptive kernel to prioritize reliable semantic predictions based on their semantic uncertainty. The kernel function in \eqref{BKI_5_reparam} is refined to incorporate uncertainty estimates of semantic predictions into the mapping process as follows: 
\begin{equation}
\begin{aligned}\label{uBKI_kernel}
    k(x_*, x_i) = 
        &\begin{cases}
            k'(d, l \cdot \beta e ^{1 - \gamma u_i}, \sigma_0) &\text{if } u_{i} \le \Uthres \\
            0 &\text{if } u_{i} > \Uthres ,
        \end{cases}
\end{aligned}
\end{equation}
where $\beta$ and $\gamma$ are hyperparameters. This uncertainty-aware kernel can enhance semantic mapping accuracy and enable more reliable calculations of the variance of map cells.

Firstly, semantic predictions with excessive uncertainty are excluded based on the uncertainty threshold $\Uthres$, which is dynamically calculated to filter out top-$\Tilde{u}$\% uncertain points. When the segmentation network lacks sufficient evidence for an input, the semantic prediction is disregarded in the mapping process. Consequently, this thresholding can prevent untrustworthy predictions from overriding accurate ones, improving the robustness against outliers in environments with perceptual uncertainties.

Then, the length of the sparse kernel is adjusted based on uncertainty to enhance the local influence of reliable predictions. The kernel length undergoes an exponential decrease in response to higher uncertainty, while the kernel length is increased for confident predictions. Increasing the kernel length for reliable predictions dynamically expands the surrounding space considered by the kernel, thereby reinforcing the local influence of reliable ones during BKI. This ensures that confident semantic predictions contribute more significantly during the recursive update of posterior parameters, enabling reliable and consistent mapping in environments with perceptual uncertainties.

\section{EXPERIMENTS}
In this section, we validate that our uncertainty-aware semantic BKI mapping framework effectively enhances semantic mapping performance in off-road settings characterized by varied visual appearances and perceptual uncertainties. Using both publicly available off-road datasets and our own dataset, we verify that our framework can improve semantic mapping performance. Then, we show that our framework can operate reliably in unseen environments using a pre-trained semantic segmentation network, facilitating robust mapping across various off-road settings. Lastly, ablation studies are performed to assess the validity of each component of our framework on its overall performance.

\subsection{Datasets}
We utilize two widely-used off-road datasets, \textit{RELLIS-3D} \cite{122_RELLIS_jiang2021rellis} and \textit{RUGD} \cite{121_RUGD_wigness2019rugd}. \textit{RELLIS-3D} contains RGB images and LiDAR point clouds with dense semantic annotations and precise robot poses, providing comprehensive information for semantic mapping. The dataset consists of $5$ sequences, and we conduct a $5$-fold evaluation where four sequences are designated for training the segmentation network and one for evaluation. \textit{RUGD} dataset provides RGB images with pixel-wise semantic annotations. As it does not provide geometric information, it is only used for network training.

To further validate our method across diverse off-road environments with varying visual appearances, we collect our dataset~(\textit{\ourdata}) using our platform equipped with an OS1-128 LiDAR and front-view RGB camera, accompanied by accurate robot poses recovered by SLAM~\cite{123_METAVerse_seo2023metaverse}. This dataset is obtained through high-speed navigation and captured under varied conditions, including different seasons, lighting conditions, and terrain types, making reliable 3D mapping more challenging. For training and evaluation, RGB images are manually annotated by experts, and the evaluation dataset contains distinct trajectory sequences not present in the training dataset. The semantic classes of all datasets are remapped to seven categories, as shown in Table~\ref{table:mainresult}.

\subsection{Experimental Setup}
\noindent\textbf{Implementation Details}
In all experiments, semantic segmentation is conducted on RGB images to leverage their richer semantic information, and the 2D semantic predictions are projected onto paired 3D point clouds. We adopt DeepLabV3~\cite{56_DeepLabV3_chen2017rethinking} with the EDL framework for semantic segmentation, and the network is trained with the loss in \eqref{EDL_OVERALL_LOSS} for $100$ epochs. For each dataset, the network is separately trained with the training subset. During the training, the hyperparameter $\lambda_{\text{kl}} = 0.5 \times ( \tau / 100 )$ is progressively increased, where $\tau$ is the current training epoch. 

\vspace{-0.05in}
\begin{table}[h]
\caption{
Hyperparameters for the semantic mapping experiments.
}
\centering
\footnotesize{
\renewcommand{\arraystretch}{1.1}
\resizebox{0.9\linewidth}{!}{%
    \begin{tabular}{c cccccc}
        \toprule
        \textbf{Dataset} & Resolution & $\alpha_0^c$ & $l$ & $\beta$ & $\gamma$ & $\Tilde{u}$ \\
        \midrule
        \textbf{\textit{RELLIS-3D}} & 0.2m & 1.0 & 0.2 & 0.5 & 10.0 & 10\% \\
        \textbf{\textit{OffRoad}}   & 0.3m & 1.0 & 0.3 & 1.0 & 3.0  & 30\% \\
        \bottomrule
    \end{tabular}
    }
    }
\label{table:hyperparams}
\end{table}

The mapping hyperparameters are summarized in \Tref{table:hyperparams}. In both datasets, the kernel scaling parameter $\sigma_0$ is set to $1.0$, to ensure the validity of the BKI framework~\cite{52_BKI_vega2014nonparametric}. To evaluate the performance of semantic mapping techniques, we first generate 3D ground truths by projecting 2D semantic labels onto a 3D point cloud. Evaluations are then conducted by querying the semantic map for each point location of the 3D ground truth and assessing the accuracy of the obtained semantic labels.

\noindent\textbf{Evaluation Metrics} Intersection over Union~($\mathrm{IoU}$) metric is employed to evaluate the semantic mapping performance. The $\mathrm{IoU}$ for the $c$-th class is computed as:
\begin{align}
    \mathrm{IoU}_c = \frac{\tpk}{\tpk+\fpk+\fnk},
\end{align} 
where $\tpk, \fpk$, and $\fnk$, represent the number of true positive, false positive, and false negative queries from the semantic map, respectively. The calculation is restricted only to queries on occupied cells, excluding any queries where the semantic map identifies them as unoccupied. Then, the mean $\mathrm{IoU}$~($\mathrm{mIoU}$) is calculated by averaging the $\mathrm{IoU}$ scores for all classes to evaluate the overall performance. Moreover, we also report voxel semantic accuracy to measure the collective accuracy of queries, which is defined as:
\begin{align}
    \mathrm{Acc} = \frac{\sum_{c=1}^{C} \tpk}{\sum_{c=1}^{C} (\tpk + \fpk)}.
\end{align}

\noindent\textbf{Comparison Methods} We perform comparative analyses with existing 3D semantic mapping methods. \textit{S-CSM} and \textit{S-BKI}~\cite{51_S-BKI_gan2020bayesian} are adopted as baselines for discrete and continuous semantic mapping, respectively. We also include results from \textit{ConvBKI}~\cite{47_ConvBKI2_wilson2023convbki}, which utilizes a learnable class-wise kernel. The learned kernel is derived from a pre-trained weight trained with \textit{RELLIS-3D}. Lastly, we compare the results of \textit{SEE-CSOM}~\cite{88_SEE-CSOM_deng2023see}, which incorporates a simple class entropy during the mapping process. 

\subsection{Quantitative and Qualitative Results}

\begin{table}[t!]
\centering
\renewcommand {\arraystretch}{1.3}
\caption{Quantitative results on \textit{RELLIS-3D} and our \textit{OffRoad} dataset. Semantic classes not present in the evaluation dataset are excluded when calculating $\mathrm{mIoU}$ and denoted with a dash (-). Our method shows superior performance compared to other 3D continuous semantic mapping methods.}
\label{tab:rellis_result}
\Large{
\resizebox{1.0\linewidth}{!}{
    \begin{tabular}{c | c | ccccccc | cc}
    \toprule
    \textbf{Dataset}
    & \textbf{Method}
    & \rotatebox{90}{\semcolor[puddle]     \hspace{0pt}puddle}
    & \rotatebox{90}{\semcolor[object]     \hspace{0pt}object}
    & \rotatebox{90}{\semcolor[paved]      \hspace{0pt}paved}
    & \rotatebox{90}{\semcolor[unpaved]    \hspace{0pt}unpaved}
    & \rotatebox{90}{\semcolor[dirt]       \hspace{0pt}dirt}
    & \rotatebox{90}{\semcolor[grass]      \hspace{0pt}grass}
    & \rotatebox{90}{\semcolor[vegetation] \hspace{0pt}vegetation}
    & \rotatebox{90}{$\mathrm{mIoU}$ [\%]}
    & \rotatebox{90}{$\mathrm{Acc}$ [\%]} \\
    \midrule \midrule
\multirow{5}{*}{\shortstack{\textbf{\textit{RELLIS}} \\ \textbf{\textit{3D}}}} &
    \textit{S-CSM~\cite{51_S-BKI_gan2020bayesian}} &
    29.6 & \textbf{34.9} & 52.9 & - & \textbf{18.4} & \textbf{77.2} & \underline{73.9} & \textbf{47.8} & \textbf{84.3} \\
    \cdashline{2-11}
    & \textit{S-BKI~\cite{51_S-BKI_gan2020bayesian}}&
    27.0 & 28.5 & \underline{53.8} & - & \underline{13.4} & 76.4 & 73.8 & 45.5 & 83.7 \\
    
    & \textit{ConvBKI~\cite{47_ConvBKI2_wilson2023convbki}} &
    \textbf{30.6} & 22.4 & 45.5 & - & 6.3 & 69.6 & 68.4 & 40.5 & 79.7 \\
    
    & \textit{SEE-CSOM~\cite{88_SEE-CSOM_deng2023see}} &
    27.7 & 33.4 & 53.4 & - & 6.7 & 76.4 & \textbf{74.0} & 45.3 & 83.6 \\
    
    & \textit{Ours} &
    \underline{29.7} & \underline{34.6} & \textbf{55.8} & - & 12.1 & \underline{76.7} & 73.8 & \underline{47.1} & \underline{84.0} \\
    \hline
\multirow{5}{*}{\textit{\ourdatabf}} &
    \textit{S-CSM~\cite{51_S-BKI_gan2020bayesian}} &
    - & \underline{43.2} & - & 72.3 & - & 52.3 & 91.5 & \underline{64.8} & 89.7 \\
    \cdashline{2-11}
    & \textit{S-BKI~\cite{51_S-BKI_gan2020bayesian}}&
    - & 38.0 & - & 68.6 & - & 49.9 & 90.9 & 61.9 & 89.0 \\
    
    & \textit{ConvBKI~\cite{47_ConvBKI2_wilson2023convbki}} &
    - & 34.7 & - & 61.2 & - & 36.8 & 91.3 & 56.0 & 88.2 \\
    
    & \textit{SEE-CSOM~\cite{88_SEE-CSOM_deng2023see}} &
    - & 41.4 & - & \underline{72.3} & - & \underline{52.3} & \underline{91.8} & 64.4 & \underline{90.1} \\
    
    & \textit{Ours} &
    - & \textbf{44.2} & - & \textbf{76.4} & - & \textbf{57.9} & \textbf{92.5} & \textbf{67.8} & \textbf{91.3} \\
    \bottomrule
    \end{tabular}
    }
}
\label{table:mainresult}
\vspace{-0.1in}
\end{table}

\begin{figure*}[t]
\begin{center}
\includegraphics[width=1.0\linewidth]{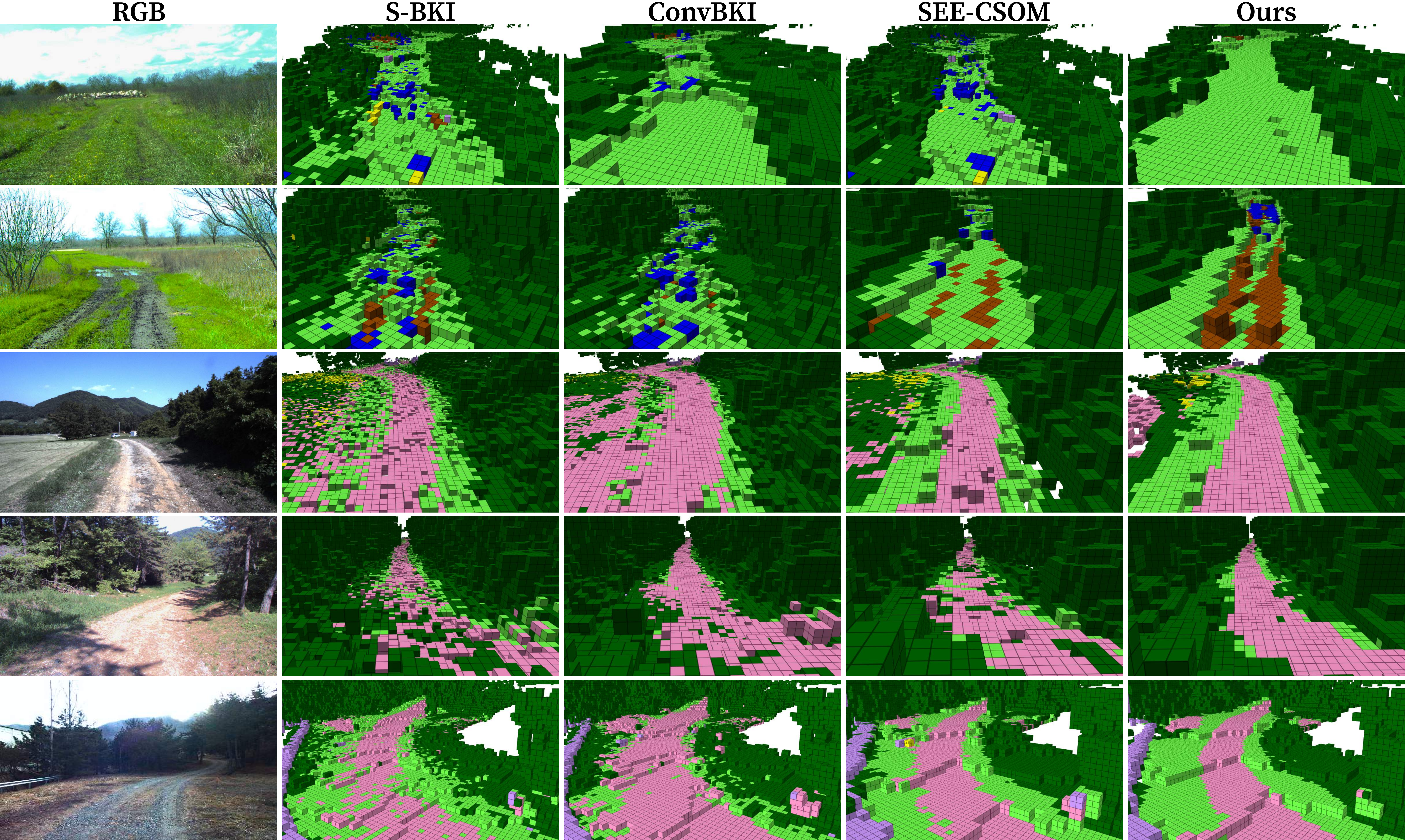}
\end{center}
\caption{Qualitative results of 3D semantic mapping methods. Compared to others, our method generates reliable and accurate maps that preserve semantic details while excluding noisy predictions. In \textit{RELLIS-3D}, only our method reconstructs \semcolorbf{grass}{grass} consistently (First row), and \semcolorbf{dirt}{dirt roads} and \semcolorbf{puddle}{puddles} in detail (Second row). In our \textit{OffRoad} dataset, our method accurately reconstructs the boundaries of \semcolorbf{unpaved}{unpaved roads}, \semcolorbf{grass}{grass}, and \semcolorbf{vegetation}{vegetation}, compared to others.
}
\label{fig:mainqualitative}
\vspace{-0.1in}
\end{figure*}

The quantitative results with \textit{RELLIS-3D} and \textit{OffRoad} are presented in \Tref{table:mainresult}. Our method shows superior performance among continuous mapping methods in both datasets. While the discrete semantic mapping approach, \textit{S-CSM}, exhibits better performance in \textit{RELLIS-3D}, this is primarily because the evaluation is conducted only on occupied voxels from discontinuous maps. In our dataset, characterized by a higher level of sparsity and perceptual uncertainty, our method surpasses \textit{S-CSM} while also constructing a continuous map. Methods adopting label inconsistency during the mapping (\textit{SEE-CSOM}) lead to enhanced performance in off-road datasets since otherwise inconsistent predictions can be negatively propagated during BKI. Our method shows better results than \textit{SEE-CSOM}, implying the relevance of EDL for quantifying semantic uncertainty and effectively utilizing this uncertainty through the adaptive kernel. Moreover, \textit{ConvBKI} leads to even poorer performance, suggesting that pre-trained kernels for specific semantic classes would result in inaccuracies in uncertain unstructured environments.

\Fref{fig:mainqualitative} provides a qualitative comparison of 3D semantic maps generated by different methods. The first and second rows display results from \textit{RELLIS-3D}, while the subsequent rows show results from our \textit{OffRoad}. Semantic mapping methods that do not incorporate perceptual uncertainty tend to produce noisy semantic maps. Our method achieves more accurate semantic map generation by conducting uncertainty quantification and integrating this uncertainty into the mapping process.

\subsection{Robust Semantic Mapping in Off-road}
We show that our uncertainty-aware 3D semantic mapping method can be robustly deployed in various off-road settings. Given that robots operating in off-road scenes frequently encounter unseen surroundings, we simulate deployments in unseen environments. To assess the capability of our method to perform robustly in a zero-shot manner, we utilize a pre-trained semantic segmentation network trained on distinct datasets. Specifically, the semantic segmentation network is trained on four different datasets: our off-road dataset~(\textit{\ourdata}), a subset of our off-road dataset~(\textit{\ourdataP}), \textit{RUGD}, and \textit{RELLIS-3D}. Then, the performance of semantic mapping is evaluated without adaptation on our off-road dataset.

\begin{table}[t!]
\centering
\renewcommand{\arraystretch}{1.35}
\caption{Quantitative results on our \textit{OffRoad} dataset using semantic segmentation networks pre-trained on different datasets.
}
\label{tab:varioussemantic}
\large{
\resizebox{1.0\linewidth}{!}{%
    \begin{tabular}{c cc  cc  cc  cc}
        \toprule
        \rowcolor[HTML]{E6E6E6} \textbf{\textit{Training Data}} & \multicolumn{2}{c}{\textit{\ourdatabf}} & \multicolumn{2}{c}{\textit{\ourdataPbf}} & \multicolumn{2}{c}{\textbf{\textit{RUGD}}} & \multicolumn{2}{c}{\textbf{\textit{RELLIS-3D}}} \\
        \midrule \midrule
        \textbf{Method} & $\mathrm{mIoU}$ & $\mathrm{Acc}$ & $\mathrm{mIoU}$ & $\mathrm{Acc}$ & $\mathrm{mIoU}$ & $\mathrm{Acc}$ & $\mathrm{mIoU}$ & $\mathrm{Acc}$ \\
        \cmidrule(lr){1-1} \cmidrule(lr){2-3} \cmidrule(lr){4-5} \cmidrule(lr){6-7} \cmidrule(lr){8-9}
        \textit{S-CSM~\cite{51_S-BKI_gan2020bayesian}} & 
        \underline{64.8} & 89.7 & \underline{60.4} & \underline{87.0} & 41.2 & 71.1 & \underline{21.9} & 65.9 \\
        \cdashline{1-9}
        
        \textit{S-BKI~\cite{51_S-BKI_gan2020bayesian}} &
        61.9 & 89.0 & 57.7 & 86.1 & 39.4 & 71.0 & \textbf{22.0} & 65.7\\
        
        \textit{ConvBKI~\cite{47_ConvBKI2_wilson2023convbki}} &
        56.0 & 88.2 & 52.4 & 85.1 & 39.3 & 71.2 & 20.8 & 64.2 \\
        
        \textit{SEE-CSOM~\cite{88_SEE-CSOM_deng2023see}} &
        64.4 & \underline{90.1} & 59.2 & 86.9 & \textbf{42.1} & \underline{71.4} & 21.6 & \textbf{66.4} \\
        
        \textit{Ours} &
        \textbf{67.8} & \textbf{91.3} & \textbf{61.0} & \textbf{87.7} & \underline{42.0} & \textbf{73.1} & 21.8 & \textbf{66.4} \\
        \bottomrule
    \end{tabular}     
    }
}
\vspace{-0.2in}
\end{table}

The quantitative results are summarized in \Tref{tab:varioussemantic}. Our method outperforms other methods, even when utilizing segmentation models trained with different training datasets. This highlights the robustness of our approach, which maintains stable performance under variations in the training data distribution. Despite using a subset of the training data (\textit{\ourdataP}), which introduces higher uncertainty in semantic prediction, our method remains superior to others. Additionally, methods integrating label inconsistency into the mapping process~(\textit{SEE-CSOM}) exhibit robust performance. Nevertheless, our method performs better than \textit{SEE-CSOM}, suggesting the capability to more effectively mitigate untrustworthy predictions and handle predictive uncertainties by incorporating EDL and uncertainty-aware BKI. Due to the substantial domain gap with the evaluation dataset, the performance gap observed in experiments with \textit{RELLIS-3D} trained models is less evident than others.

\begin{figure*}[t!]
\begin{center}
\includegraphics[width=1.0\linewidth]{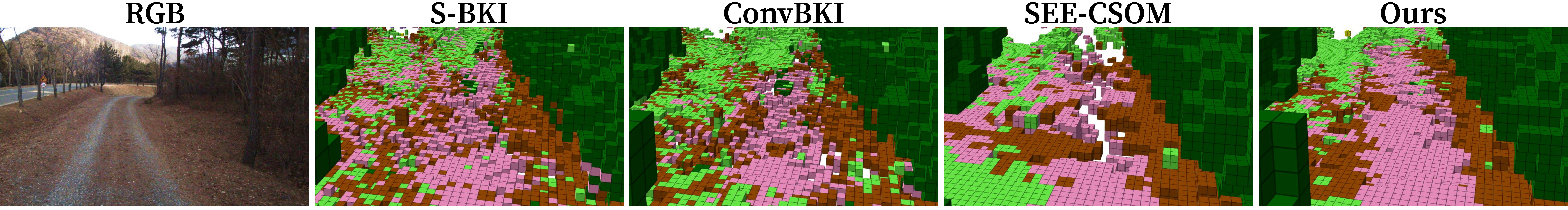}
\end{center}
\caption{Zero-shot semantic mapping results on our \textit{\ourdata} dataset using a semantic segmentation network pre-trained on \textit{RUGD}. Our method robustly constructs semantic maps despite prediction uncertainties in unseen environments, whereas other methods struggle to produce clear maps.
}
\label{fig:robustqualitative}
\vspace{-0.1in}
\end{figure*}

\Fref{fig:robustqualitative} shows the semantic mapping results using the segmentation network trained on \textit{RUGD}. While other methods struggle to reconstruct accurate semantic maps or produce discontinuous maps, our method successfully creates a clear, continuous map with unambiguous boundaries. \textit{SEE-CSOM} fails to generate fully continuous semantic maps, indicating inefficacy in reliably addressing temporal inconsistency and uncertainties in semantic predictions. Additionally, \textit{ConvBKI}, which requires dense 3D ground truths for learning the class-wise adaptive kernel, shows limited adaptability to unseen environments due to potential overfitting. In contrast, our method consistently produces continuous maps with clear boundaries without requiring additional 3D ground truth.

\subsection{Ablation Studies}

\begin{figure}[b]
\centering
\includegraphics[width=1.0\linewidth]{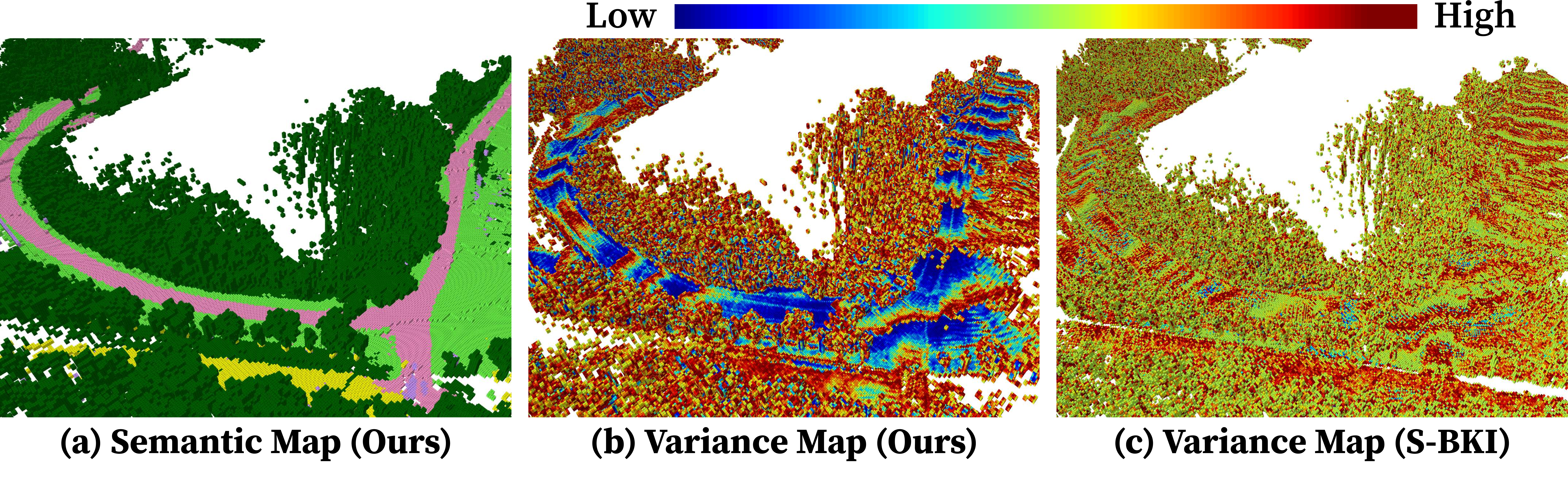}
\caption{Visualization of variance maps. Through uncertainty-aware BKI, our method generates more informative variance maps capable of reliably quantifying semantic map uncertainty.}
\label{fig:varMap}
\end{figure}

We conduct ablation studies on our uncertainty-aware BKI framework to evaluate the impact of its key components described in \Sref{METHOD_UAM}. Specifically, we examine the contribution of the Continuous Categorical extension (\textbf{Prob.})~\eqref{uBKI_final_update}, the adaptive kernel length (\textbf{Adap.})~\eqref{uBKI_kernel}, and the uncertainty thresholding (\textbf{Thr.})~\eqref{uBKI_kernel} to the overall performance.

To comprehensively evaluate the impact of each component, we further investigate Voxel Occupancy Accuracy~($\mathrm{O.Acc}$), defined as the proportion of queries with occupied voxels, measuring the geometric completeness of semantic maps. Moreover, the Brier Score~($\mathrm{BS}$) is calculated to evaluate the reliability of the variance map, which can be interpreted as representing the uncertainty of the map cells. The confidence of a map cell's semantic estimation is defined as $\mathrm{Conf}_i = 1 - \hat{\mathrm{Var}}[\alpha_t^\psi]$ by normalizing the variance in \eqref{BKI_final_variance} to $\hat{\mathrm{Var}}[\alpha_t^\psi]$. $\mathrm{BS}$ is then defined as follows:
\begin{align}
    \mathrm{BS} = \frac{1}{M} \sum^M_{i=1} (\mathbbm{1}_{\text{correct}} - \mathrm{Conf}_i)^2,
\end{align}
where $M$ is the number of occupied map cells, and the indicator function $\mathbbm{1}_{\text{correct}}$ is assigned $1$ if the semantic label of a map cell $i$ is correctly estimated as the ground truth label of the cell and $0$ otherwise. This metric becomes lower if the confidence of correctly estimated cells is high and the confidence of incorrectly estimated map cells is low.

The results are summarized in \Tref{tab:ablation}. In both datasets, the Continuous Categorical extension and the uncertainty threshold improve $\mathrm{mIoU}$, whereas the adaptive kernel length enhances the reliability of the uncertainty measure with low $\mathrm{BS}$. Although the uncertainty threshold significantly improves the mapping performances, it exacerbates geometric completeness with decreased $\mathrm{O.Acc}$. Combining the adaptive kernel length with the uncertainty threshold enhances both the semantic mapping performance and the uncertainty's reliability while maintaining geometric completeness. The variance map illustrated in \Fref{fig:varMap} shows that the variance of our map holds more meaningful uncertainty information, as quantitatively described with improved $\mathrm{BS}$.

\newcommand{\cmark}{\ding{51}}%
\begin{table}[t!]
\centering
\renewcommand{\arraystretch}{1.4}
\caption{
 Results of the ablation studies.
}
\label{tab:ablation}
\Large{
\resizebox{1.0\linewidth}{!}{%
    \begin{tabular}{ccc cccc cccc}
        \toprule
        &&&
        \multicolumn{4}{c}{\textbf{\textit{RELLIS-3D}}} & \multicolumn{4}{c}{\textit{\ourdatabf}} \\
        \cmidrule(lr){4-7} \cmidrule(lr){8-11} 
        \multirow{1}{*}{\textbf{Prob.}} & \multirow{1}{*}{\textbf{Adap.}} & \multirow{1}{*}{\textbf{Thr.}} & $\mathrm{mIoU}$ & $\mathrm{Acc}$ & $\mathrm{O.Acc}$ & $\mathrm{BS}(\downarrow)$ & $ \mathrm{mIoU}$ & $\mathrm{Acc}$ & $\mathrm{O.Acc}$ & $\mathrm{BS}(\downarrow)$ \\
        \midrule \midrule
        - & - & \multicolumn{1}{c|}{-}& 45.5 & 83.7 & 96.7 & 35.7 & 61.9 & 89.0 & 99.7 & 39.7 \\
        \cmark & - & \multicolumn{1}{c|}{-} & 46.4 & 83.6 & 96.7 & 40.1 & 63.4 & 89.5 & \textbf{99.9} & 51.8 \\
        \cmark & \cmark & \multicolumn{1}{c|}{-} & 46.3 & 83.8 & \textbf{96.9} & \underline{27.4} & 64.9 & 90.2 & \textbf{99.9} & \textbf{12.5} \\
        \cmark & - & \multicolumn{1}{c|}{\cmark} & \textbf{47.8} & \textbf{84.4} & 95.5 & 43.1 & \textbf{68.9} & \textbf{91.9} & 95.4 & 51.8 \\
        \cmark & \cmark & \multicolumn{1}{c|}{\cmark} & \underline{47.1} & \underline{84.0} & \textbf{96.9} & \textbf{24.1} & \underline{67.8} & \underline{91.3} & 98.2 & \textbf{12.5} \\
        \bottomrule
    \end{tabular}
    }
}
\vspace{-0.2in}
\end{table}

\section{CONCLUSIONS}
This paper presents an uncertainty-aware Bayesian Kernel Inference framework for reliable semantic mapping in uncertain environments. To address the challenges of high perceptual uncertainties, our framework incorporates uncertainty of semantic predictions from DNNs into the mapping process. Specifically, we leverage Evidential Deep Learning for semantic segmentation and introduce the adaptive kernel and uncertainty threshold mechanism to address the semantic uncertainty. Through extensive experiments, we demonstrate the effectiveness of our framework in handling semantic uncertainty, resulting in accurate and reliable semantic maps in perceptually challenging unstructured outdoor environments. Our method is robust in unseen environments with higher semantic uncertainty, constructing continuous semantic maps with precise semantic boundaries. We believe our framework has the potential to improve the accuracy and reliability of semantic mapping in a variety of contexts with perceptual uncertainties beyond off-road environments.

Future works include enhancing the real-time capability of our framework with more efficient geometric models. Additionally, we intend to extend the framework through uncertainty-aware multi-modal fusion, enabling effective integration of information. Furthermore, we are interested in advanced techniques for uncertainty calibration to ensure more reliable mapping. Exploring the potential of our semantic map and its uncertainty for active exploration also presents an interesting research avenue.

\addtolength{\textheight}{0cm}   
                                  
\bibliographystyle{IEEEtran}
\bibliography{mybib.bib}

\begin{thebibliography}{10}
\providecommand{\url}[1]{#1}
\csname url@samestyle\endcsname
\providecommand{\newblock}{\relax}
\providecommand{\bibinfo}[2]{#2}
\providecommand{\BIBentrySTDinterwordspacing}{\spaceskip=0pt\relax}
\providecommand{\BIBentryALTinterwordstretchfactor}{4}
\providecommand{\BIBentryALTinterwordspacing}{\spaceskip=\fontdimen2\font plus
\BIBentryALTinterwordstretchfactor\fontdimen3\font minus \fontdimen4\font\relax}
\providecommand{\BIBforeignlanguage}[2]{{%
\expandafter\ifx\csname l@#1\endcsname\relax
\typeout{** WARNING: IEEEtran.bst: No hyphenation pattern has been}%
\typeout{** loaded for the language `#1'. Using the pattern for}%
\typeout{** the default language instead.}%
\else
\language=\csname l@#1\endcsname
\fi
#2}}
\providecommand{\BIBdecl}{\relax}
\BIBdecl

\bibitem{131_OccupancyMap_elfes1989using}
A.~Elfes, ``Using occupancy grids for mobile robot perception and navigation,'' \emph{Computer}, vol.~22, no.~6, pp. 46--57, 1989.

\bibitem{6_OctoMap_hornung2013octomap}
A.~Hornung, K.~M. Wurm, M.~Bennewitz, C.~Stachniss, and W.~Burgard, ``Octomap: An efficient probabilistic 3d mapping framework based on octrees,'' \emph{Autonomous Robots}, vol.~34, pp. 189--206, 2013.

\bibitem{50_GPOctoMap_wang2016fast}
J.~Wang and B.~Englot, ``Fast, accurate gaussian process occupancy maps via test-data octrees and nested bayesian fusion,'' in \emph{IEEE International Conference on Robotics and Automation (ICRA)}, 2016, pp. 1003--1010.

\bibitem{49_BGKOctoMap_doherty2017bayesian}
K.~Doherty, J.~Wang, and B.~Englot, ``Bayesian generalized kernel inference for occupancy map prediction,'' in \emph{IEEE International Conference on Robotics and Automation (ICRA)}, 2017, pp. 3118--3124.

\bibitem{52_BKI_vega2014nonparametric}
W.~R. Vega-Brown, M.~Doniec, and N.~G. Roy, ``Nonparametric bayesian inference on multivariate exponential families,'' \emph{Advances in Neural Information Processing Systems (NeurIPS)}, vol.~27, 2014.

\bibitem{94_kim20133d}
B.-s. Kim, P.~Kohli, and S.~Savarese, ``3d scene understanding by voxel-crf,'' in \emph{IEEE/CVF International Conference on Computer Vision (CVPR)}, 2013, pp. 1425--1432.

\bibitem{103_valentin2013mesh}
J.~P. Valentin, S.~Sengupta, J.~Warrell, A.~Shahrokni, and P.~H. Torr, ``Mesh based semantic modelling for indoor and outdoor scenes,'' in \emph{IEEE/CVF Conference on Computer Vision and Pattern Recognition (CVPR)}, 2013, pp. 2067--2074.

\bibitem{95_SemanticOctree_sengupta2015semantic}
S.~Sengupta and P.~Sturgess, ``Semantic octree: Unifying recognition, reconstruction and representation via an octree constrained higher order mrf,'' in \emph{IEEE International Conference on Robotics and Automation (ICRA)}, 2015, pp. 1874--1879.

\bibitem{109_paz2020probabilistic}
D.~Paz, H.~Zhang, Q.~Li, H.~Xiang, and H.~I. Christensen, ``Probabilistic semantic mapping for urban autonomous driving applications,'' in \emph{IEEE/RSJ International Conference on Intelligent Robots and Systems (IROS)}, 2020, pp. 2059--2064.

\bibitem{23_SSMI_asgharivaskasi2023semantic}
A.~Asgharivaskasi and N.~Atanasov, ``Semantic octree mapping and shannon mutual information computation for robot exploration,'' \emph{IEEE Transactions on Robotics}, 2023.

\bibitem{26_morilla2023robust}
D.~Morilla-Cabello, L.~Mur-Labadia, R.~Martinez-Cantin, and E.~Montijano, ``Robust fusion for bayesian semantic mapping,'' in \emph{IEEE/RSJ International Conference on Intelligent Robots and Systems (IROS)}, 2023.

\bibitem{106_FusionOverconfidence_marques2023overconfidence}
J.~M.~C. Marques, A.~Zhai, S.~Wang, and K.~Hauser, ``On the overconfidence problem in semantic 3d mapping,'' \emph{arXiv preprint arXiv:2311.10018}, 2023.

\bibitem{51_S-BKI_gan2020bayesian}
L.~Gan, R.~Zhang, J.~W. Grizzle, R.~M. Eustice, and M.~Ghaffari, ``Bayesian spatial kernel smoothing for scalable dense semantic mapping,'' \emph{IEEE Robotics and Automation Letters}, vol.~5, no.~2, pp. 790--797, 2020.

\bibitem{96_SemanticFusion_mccormac2017semanticfusion}
J.~McCormac, A.~Handa, A.~Davison, and S.~Leutenegger, ``Semanticfusion: Dense 3d semantic mapping with convolutional neural networks,'' in \emph{IEEE International Conference on Robotics and automation (ICRA)}, 2017, pp. 4628--4635.

\bibitem{97_yang2017semantic}
S.~Yang, Y.~Huang, and S.~Scherer, ``Semantic 3d occupancy mapping through efficient high order crfs,'' in \emph{IEEE/RSJ International Conference on Intelligent Robots and Systems (IROS)}, 2017, pp. 590--597.

\bibitem{98_DA-RNN_xiang2017rnn}
Y.~Xiang and D.~Fox, ``Da-rnn: Semantic mapping with data associated recurrent neural networks,'' \emph{Robotics: Science and Systems (RSS)}, 2017.

\bibitem{99_RecurrentOctoMap_sun2018recurrent}
L.~Sun, Z.~Yan, A.~Zaganidis, C.~Zhao, and T.~Duckett, ``Recurrent-octomap: Learning state-based map refinement for long-term semantic mapping with 3-d-lidar data,'' \emph{IEEE Robotics and Automation Letters}, vol.~3, no.~4, pp. 3749--3756, 2018.

\bibitem{105_maturana2018real}
D.~Maturana, P.-W. Chou, M.~Uenoyama, and S.~Scherer, ``Real-time semantic mapping for autonomous off-road navigation,'' in \emph{Field and Service Robotics: Results of the 11th International Conference}, 2018, pp. 335--350.

\bibitem{107_guo2017calibration}
C.~Guo, G.~Pleiss, Y.~Sun, and K.~Q. Weinberger, ``On calibration of modern neural networks,'' in \emph{International Conference on Machine Learning (ICML)}, 2017, pp. 1321--1330.

\bibitem{83_ModernReliability_de2023reliability}
P.~de~Jorge, R.~Volpi, P.~H. Torr, and G.~Rogez, ``Reliability in semantic segmentation: Are we on the right track?'' in \emph{IEEE/CVF Conference on Computer Vision and Pattern Recognition (CVPR)}, 2023, pp. 7173--7182.

\bibitem{86_OFFROAD_jin2021memory}
Y.~Jin, D.~Han, and H.~Ko, ``Memory-based semantic segmentation for off-road unstructured natural environments,'' in \emph{IEEE/RSJ International Conference on Intelligent Robots and Systems (IROS)}, 2021, pp. 24--31.

\bibitem{seo2023learning}
J.~Seo, S.~Sim, and I.~Shim, ``Learning off-road terrain traversability with self-supervisions only,'' \emph{IEEE Robotics and Automation Letters}, 2023.

\bibitem{88_SEE-CSOM_deng2023see}
Y.~Deng, M.~Wang, Y.~Yang, D.~Wang, and Y.~Yue, ``See-csom: Sharp-edged and efficient continuous semantic occupancy mapping for mobile robots,'' \emph{IEEE Transactions on Industrial Electronics}, 2023.

\bibitem{47_ConvBKI2_wilson2023convbki}
J.~Wilson, Y.~Fu, J.~Friesen, P.~Ewen, A.~Capodieci, P.~Jayakumar, K.~Barton, and M.~Ghaffari, ``Convbki: Real-time probabilistic semantic mapping network with quantifiable uncertainty,'' \emph{arXiv preprint arXiv:2310.16020}, 2023.

\bibitem{10_EDL_sensoy2018evidential}
M.~Sensoy, L.~Kaplan, and M.~Kandemir, ``Evidential deep learning to quantify classification uncertainty,'' \emph{Advances in Neural Information Processing Systems (NeurIPS)}, vol.~31, 2018.

\bibitem{4_GMMap_li2024gmmap}
P.~Z.~X. Li, S.~Karaman, and V.~Sze, ``Gmmap: Memory-efficient continuous occupancy map using gaussian mixture model,'' \emph{IEEE Transactions on Robotics}, 2024.

\bibitem{92_sengupta2013urban}
S.~Sengupta, E.~Greveson, A.~Shahrokni, and P.~H. Torr, ``Urban 3d semantic modelling using stereo vision,'' in \emph{IEEE International Conference on Robotics and Automation (ICRA)}, 2013, pp. 580--585.

\bibitem{101_kundu2014joint}
A.~Kundu, Y.~Li, F.~Dellaert, F.~Li, and J.~M. Rehg, ``Joint semantic segmentation and 3d reconstruction from monocular video,'' in \emph{European Conference on Computer Vision (ECCV)}, 2014, pp. 703--718.

\bibitem{104_vineet2015incremental}
V.~Vineet, O.~Miksik, M.~Lidegaard, M.~Nie{\ss}ner, S.~Golodetz, V.~A. Prisacariu, O.~K{\"a}hler, D.~W. Murray, S.~Izadi, P.~P{\'e}rez \emph{et~al.}, ``Incremental dense semantic stereo fusion for large-scale semantic scene reconstruction,'' in \emph{IEEE International Conference on Robotics and Automation (ICRA)}, 2015, pp. 75--82.

\bibitem{118_BNNs_jospin2022hands}
L.~V. Jospin, H.~Laga, F.~Boussaid, W.~Buntine, and M.~Bennamoun, ``Hands-on bayesian neural networks—a tutorial for deep learning users,'' \emph{IEEE Computational Intelligence Magazine}, vol.~17, no.~2, pp. 29--48, 2022.

\bibitem{117_MCDropout_gal2016dropout}
Y.~Gal and Z.~Ghahramani, ``Dropout as a bayesian approximation: Representing model uncertainty in deep learning,'' in \emph{International Conference on Machine Learning (ICML)}, 2016, pp. 1050--1059.

\bibitem{110_DeepEnsemble_lakshminarayanan2017simple}
B.~Lakshminarayanan, A.~Pritzel, and C.~Blundell, ``Simple and scalable predictive uncertainty estimation using deep ensembles,'' \emph{Advances in Neural Information Processing Systems (NeurIPS)}, vol.~30, 2017.

\bibitem{kim2023bridging}
T.~Kim, J.~Mun, J.~Seo, B.~Kim, and S.~Hong, ``Bridging active exploration and uncertainty-aware deployment using probabilistic ensemble neural network dynamics,'' \emph{Robotics: Science and Systems (RSS)}, 2023.

\bibitem{129_DST_yager2008classic}
R.~R. Yager and L.~Liu, \emph{Classic works of the Dempster-Shafer theory of belief functions}.\hskip 1em plus 0.5em minus 0.4em\relax Springer, 2008, vol. 219.

\bibitem{70_USNet_chang2022fast}
Y.~Chang, F.~Xue, F.~Sheng, W.~Liang, and A.~Ming, ``Fast road segmentation via uncertainty-aware symmetric network,'' in \emph{IEEE International Conference on Robotics and Automation (ICRA)}, 2022, pp. 11\,124--11\,130.

\bibitem{72_TMCJournal_han2022trusted}
Z.~Han, C.~Zhang, H.~Fu, and J.~T. Zhou, ``Trusted multi-view classification with dynamic evidential fusion,'' \emph{IEEE Transactions on Pattern Analysis and Machine Intelligence}, vol.~45, no.~2, pp. 2551--2566, 2022.

\bibitem{13_EvPSNet_sirohi2023uncertainty}
K.~Sirohi, S.~Marvi, D.~B{\"u}scher, and W.~Burgard, ``Uncertainty-aware panoptic segmentation,'' \emph{IEEE Robotics and Automation Letters}, vol.~8, no.~5, pp. 2629--2636, 2023.

\bibitem{120_EVORA_cai2023evora}
X.~Cai, S.~Ancha, L.~Sharma, P.~R. Osteen, B.~Bucher, S.~Phillips, J.~Wang, M.~Everett, N.~Roy, and J.~P. How, ``Evora: Deep evidential traversability learning for risk-aware off-road autonomy,'' \emph{arXiv preprint arXiv:2311.06234}, 2023.

\bibitem{66_SparseKernel_melkumyan2009sparse}
A.~Melkumyan and F.~T. Ramos, ``A sparse covariance function for exact gaussian process inference in large datasets,'' in \emph{International Joint Conference on Artificial Intelligence (IJCAI)}, 2009.

\bibitem{74_RED_pandey2023learn}
D.~S. Pandey and Q.~Yu, ``Learn to accumulate evidence from all training samples: theory and practice,'' in \emph{International Conference on Machine Learning (ICML)}, 2023, pp. 26\,963--26\,989.

\bibitem{89_gordon2020continuous}
E.~Gordon-Rodriguez, G.~Loaiza-Ganem, and J.~Cunningham, ``The continuous categorical: a novel simplex-valued exponential family,'' in \emph{International Conference on Machine Learning (ICML)}, 2020, pp. 3637--3647.

\bibitem{122_RELLIS_jiang2021rellis}
P.~Jiang, P.~Osteen, M.~Wigness, and S.~Saripalli, ``Rellis-3d dataset: Data, benchmarks and analysis,'' in \emph{IEEE International Conference on Robotics and Automation (ICRA)}, 2021, pp. 1110--1116.

\bibitem{121_RUGD_wigness2019rugd}
M.~Wigness, S.~Eum, J.~G. Rogers, D.~Han, and H.~Kwon, ``A rugd dataset for autonomous navigation and visual perception in unstructured outdoor environments,'' in \emph{IEEE/RSJ International Conference on Intelligent Robots and Systems (IROS)}, 2019, pp. 5000--5007.

\bibitem{123_METAVerse_seo2023metaverse}
J.~Seo, T.~Kim, S.~Ahn, and K.~Kwak, ``Metaverse: Meta-learning traversability cost map for off-road navigation,'' \emph{arXiv preprint arXiv:2307.13991}, 2023.

\bibitem{56_DeepLabV3_chen2017rethinking}
L.-C. Chen, G.~Papandreou, F.~Schroff, and H.~Adam, ``Rethinking atrous convolution for semantic image segmentation,'' \emph{arXiv preprint arXiv:1706.05587}, 2017.

\end{thebibliography}
\end{document}